\documentclass[conference]{IEEEtran}
\IEEEoverridecommandlockouts
\usepackage{cite}
\usepackage{amsmath,amssymb,amsfonts}
\usepackage{algorithmic}
\usepackage{graphicx}
\usepackage{textcomp}
\usepackage{url}
\usepackage{comment}

\usepackage{listings}
\usepackage{xcolor}
\def\BibTeX{{\rm B\kern-.05em{\sc i\kern-.025em b}\kern-.08em
    T\kern-.1667em\lower.7ex\hbox{E}\kern-.125emX}}
\begin{document}

\title{Towards Secure Cloud-Native Computing: Unveiling Kubernetes Misconfigurations with Large Language Models\\
}

\author{\IEEEauthorblockN{1\textsuperscript{st} Mostafa Anouar Ghorab}
\IEEEauthorblockA{\textit{Department of Computer Science and Software Engineering} \\
\textit{Laval University}\\
Quebec City, Canada  \\
mostafa-anouar.ghorab.1@ulaval.ca}
\and
\IEEEauthorblockN{2\textsuperscript{nd} Mohamed Aymen Saied}
\IEEEauthorblockA{\textit{Department of Computer Science and Software Engineering} \\
\textit{Laval University}\\
Quebec City, Canada  \\
mohamed-aymen.saied@ift.ulaval.ca}

}
\maketitle

\begin{abstract}
In the rapidly evolving landscape of cloud-native computing, Organizations are increasingly adopting infrastructure models that emphasize scalability, flexibility, and efficiency. Kubernetes has become the de facto standard for orchestrating containerized applications in these environments. However, the inherent complexity of cloud-native ecosystems introduces significant challenges, particularly in the form of misconfigurations that can compromise both security and performance.

This study explores the potential of Large Language Models (LLMs) in identifying Kubernetes misconfigurations. We introduce a comprehensive taxonomy of common misconfiguration types, offering a structured framework to better understand and categorize these issues. Additionally, we conduct an empirical evaluation of state-of-the-art detection tools to benchmark their effectiveness. Furthermore, we analyze the Kubernetes objects most prone to misconfiguration and evaluate the severity of the identified issues. By leveraging advanced machine learning techniques, including LLMs, we provide novel insights into enhancing misconfiguration detection methodologies.
\end{abstract}


\maketitle
\vspace{-0.5em}
\section{Introduction}
Kubernetes has rapidly become the de facto standard for container orchestration in cloud-native systems, empowering organizations to deploy, manage, and scale applications seamlessly across dynamic and distributed environments  \cite{brun}. However, as with any complex system, the power and flexibility of Kubernetes come with significant challenges, particularly in the realm of configuration management. Kubernetes misconfigurations have become a major concern, often leading to performance degradation, security vulnerabilities, and even system outages \cite{b35}.

As organizations are increasingly adopting Kubernetes to manage complex workloads and support agile development processes, the risk of human error rises in parallel with the growing complexity of configurable parameters \cite{b40}. Recent statistics indicate an alarming increase in security breaches due to Kubernetes misconfigurations \cite{b28}.

To mitigate this challenge, various tools and methodologies such as Kube Score \cite{b10}, Snyk \cite{b9}, Datree \cite{b8}, and others \cite{b4} \cite{b7} \cite{b11} have been developed to detect misconfigurations. However, these tools generally operate in isolation, relying on proprietary classifications without a common reference framework. At present, misconfigurations are identified primarily through tool-specific names and descriptions, leading to inconsistencies and a fragmented, non-standardized approach. This absence of a unified taxonomy undermines the efficiency of detection and remediation efforts, making it challenging to recognize recurring misconfiguration patterns and assess their associated risks.

To address these pressing challenges, this research paper aims to provide a comprehensive examination of Kubernetes misconfigurations through the lens of three central research questions. First, we investigate methodologies for developing a comprehensive taxonomy of Kubernetes misconfigurations, identifying key categories that will enhance the understanding and classification of these issues (RQ1). This taxonomy is intended to serve as a foundational framework for standardizing the detection and management of misconfigurations across diverse Kubernetes environments.

Secondly, we analyze the frequency of Kubernetes misconfigurations across different object types, assign a corresponding risk level to each type, and assess the effectiveness of LLMs in detecting these misconfigurations (RQ2). As the adoption of machine learning for automation continues to grow, this analysis offers valuable insights into both the capabilities and limitations of LLMs in the context of Kubernetes configuration management.

Finally, we compare the results of our best-performing LLM-based model with those of existing misconfiguration detection tools to assess how our proposed approach performs relative to state-of-the-art methods (RQ3). This comparative analysis aims to illuminate the strengths and weaknesses of current tools while providing insights into areas for improvement. By understanding how our approach aligns with existing solutions, we aspire to contribute to the development of more effective misconfiguration detection strategies.

The remainder of this paper is organized as follows: Section 2 reviews related work on Kubernetes misconfigurations, focusing on existing taxonomies and detection tools. Section 3 presents the proposed methodology, detailing the unified taxonomy and the application of LLMs for detection. Section 4 describes the evaluation protocol, including experimental setup, metrics, results, and a comparison with established tools. Section 5 addresses threats to validity and study limitations. Finally, Section 6 concludes and suggests future research directions to enhance Kubernetes misconfiguration detection and management. The replication package is available at: https://anonymous.4open.science/r/138-CLOUD2025-77EC/readme.md

\vspace{-0.5em}
\section{Related work}

Detecting Kubernetes misconfigurations is essential for ensuring the security and reliability of containerized applications. Previous research has explored various methodologies, including rule-based detection, empirical studies, and machine learning based approaches.

Felix Klement et al. \cite{b2} conducted a comprehensive investigation into vulnerabilities and misconfigurations within Kubernetes-based intelligent controller clusters in Open Radio Access Network infrastructures. Their study highlighted security risks arising from outdated software packages and assessed the severity of these vulnerabilities. To mitigate these risks, they proposed integrating security evaluation mechanisms into the deployment process, ensuring that the latest software versions are utilized to enhance system security and reduce potential threats. Another proposed approach involved a centralized logging solution that systematically aggregates configuration files, applies multiple misconfiguration detection tools for in-depth analysis, eliminates redundant misconfigurations, and generates a detailed diagnostic report. This report is presented through a web interface, offering real-time visualization of the operating state of the clusterter and highlighting detected misconfigurations and security issues \cite{b3}.

Ehud Malul et al. \cite{b4} introduced GenKubeSec, a method leveraging LLMs to detect and identify Kubernetes misconfigurations. While GenKubeSec demonstrates high recall and achieves comparable precision to rule-based tools, its performance is constrained by the size of configuration files, successfully analyzing only 82\% of the collected configurations. Another adaptive approach involves an autonomous controller designed to detect and mitigate privilege escalation misconfigurations in Kubernetes clusters and IoT devices. Upon identifying an exploited privilege escalation, the controller autonomously executes a predefined remediation strategy, which may include terminating the affected process, redeploying compromised components, or performing a controlled restart to restore a secure state \cite{b5}.

An empirical analysis of 2,039 Kubernetes manifests identified eleven distinct types of security misconfigurations. To systematically detect these misconfigurations, researchers defined a set of detection rules and developed SLI-KUBE, a tool designed to implement these rules and quantify the prevalence of security issues \cite{b6}. In a related contribution, Francesco Minna et al. \cite{b7} proposed an analysis pipeline for Helm charts, evaluating the effectiveness of rule-based tools such as Checkov and KICS, while also assessing the potential of LLMs for mitigating misconfigurations.

Beyond academic research, several industry tools have been developed to detect Kubernetes misconfigurations. Datree \cite{b8} ensures the compliance of policies in CI / CD pipelines by validating the configurations against predefined rules. Snyk \cite{b9} combines rule-based analysis with known CVEs to detect security vulnerabilities and misconfigurations in Kubernetes manifests. KubeScore \cite{b10} performs static analysis of Kubernetes configurations and provides recommendations based on best practices. Despite their effectiveness,all rule-based approaches including Checkov \cite{b11}, KICS \cite{b12}, and Polaris \cite{b24, b25} have inherent limitations. They rely on static rule sets, struggle to detect novel misconfigurations, and often produce false positives or negatives. Furthermore, as Kubernetes environments grow in complexity, these tools face challenges related to scalability and contextual awareness, limiting their ability to adapt to evolving configurations.

To overcome these limitations, emerging methodologies leverage machine learning and LLMs to detect vulnerabilities and misconfigurations. Unlike static rule-based tools, ML-driven approaches adapt to new data and enhance contextual understanding, enabling a more precise identification of misconfigurations. Several studies \cite{b35, b36, b38, b39} have demonstrated the potential of LLMs to detect anomalies in various ecosystems, however, \cite{b4} is the only work that investigates the potential of LLMs in detecting Kubernetes misconfiguration.

Our research differentiates itself from both rule-based systems and machine learning-based tools in several key aspects. Unlike Datree, Snyk, KubeScore, Checkov, KICS, and Polaris, which rely on predefined rule sets, our study explores the adaptability of LLMs to detect misconfigurations within their deployment context. Rule-based tools, while effective for identifying known misconfigurations, are constrained by their static nature and require continuous manual updates to address emerging vulnerabilities. In contrast, LLMs can dynamically interpret and generalize from diverse configuration patterns, enabling the identification of novel and context-dependent misconfigurations beyond predefined rules. This adaptability enhances detection accuracy, particularly for newly emerging misconfigurations that may not yet be captured by traditional rule-based systems.

Moreover, while existing rule-based tools derive their detection rules from sources such as CVEs, security best practices, and compliance guidelines, their effectiveness is limited by the completeness and timeliness of these rules. Our research addresses this challenge by using LLMs trained on the detection outcomes of multiple rule-based tools, allowing the model to learn from a broader and more diverse set of misconfiguration patterns. This approach enhances misconfiguration detection rates and broadens pattern coverage, thereby improving the overall robustness of the detection process.

Another methodological distinction lies in data segmentation. Unlike previous work such as GenKubeSec, which processes entire configuration files as single units and fails to analyze larger configurations due to token limitations (processing only 82\% of collected configurations), our method effectively segments files into smaller components. This ensures that we can analyze configurations of any size without restrictions, leading to comprehensive misconfiguration detection. As a result, our method not only surpasses previous techniques in coverage but also delivers superior accuracy, scalability, and robustness in identifying misconfigurations across diverse Kubernetes configurations.
\vspace{-0.5em}
\section{Proposed Approach} \label{sec:DataCollection}
Kubernetes is a pivotal component in modern native cloud environments, facilitating the deployment, scaling, and operation of containerized applications\cite{STIRBU2024107529}. However, its inherent complexity and extensive configuration options render it vulnerable to misconfigurations, which can compromise system security and reliability. 

This research aims to enhance security in native cloud environments by proposing four key contributions.

\begin{itemize}

    \item Development of a Misconfiguration Taxonomy: A systematic methodology is proposed to construct a comprehensive taxonomy of Kubernetes misconfigurations. This taxonomy categorizes and classifies common misconfigurations, providing a structured framework for analyzing, prioritizing and mitigating configuration issues.
    
    \item Identification of Susceptible Kubernetes Objects : The study identifies Kubernetes objects most prone to misconfigurations, with a focus on those that pose the highest security and operational risks. 
    
    \item Application of LLMs for Misconfiguration Detection: The potential of LLMs for detecting misconfigurations is investigated through the development of four distinct datasets and the application of multiple models to identify anomalous and potentially dangerous configurations.
    
    \item Comparative analysis of detection tools : A systematic evaluation of existing misconfiguration detection tools is conducted, assessing their coverage, detection accuracy, and overall effectiveness. This analysis highlights their respective strengths and limitations, providing insights into potential areas for improvement.

\end{itemize}

\begin{figure*}[h!]
\centering
\includegraphics[width=13cm, height=8cm]{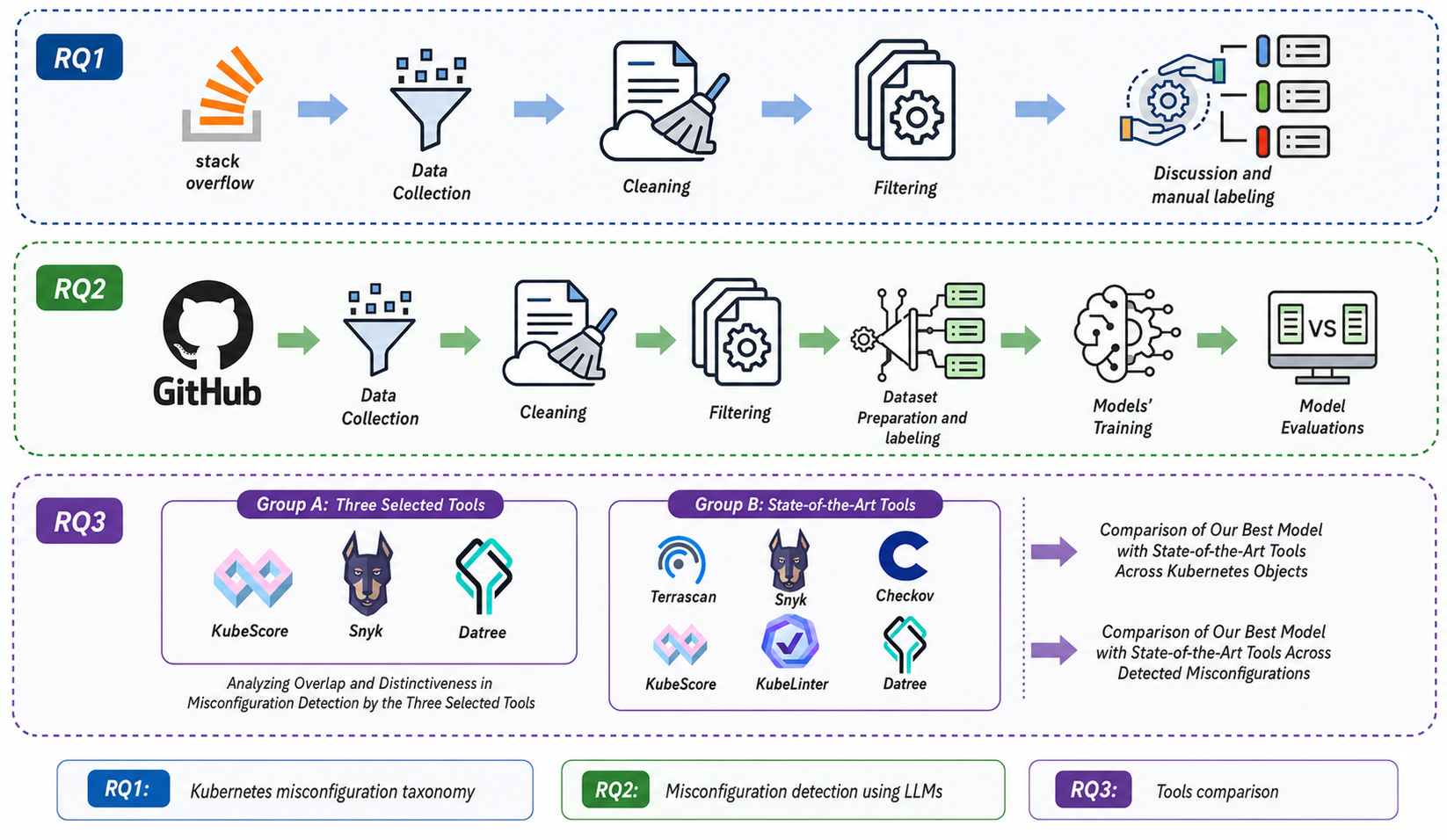} 
\caption{Overview of our study’s methodology}
\label{fig:Approche}
\end{figure*}

Figure \ref{fig:Approche} provides a comprehensive overview of our proposed methodology.
For the first research question, the process begins with the collection of posts from Stack Overflow, progressing through the development of the final taxonomy.
In addressing the second research question, we start by collecting Kubernetes configuration files from GitHub, followed by an analysis of the most susceptible Kubernetes components. We then identify which of these objects pose the highest security risks based on the severity of their misconfigurations. Subsequently, we construct four distinct datasets, train the selected models on them, and evaluate their performance.
Finally, for the third research question, we conducted a comparative analysis by benchmarking the performance of our best-performing model against existing tools.

\begin{figure}[h!]
\centering
\includegraphics[width=8cm, height=4.3cm]{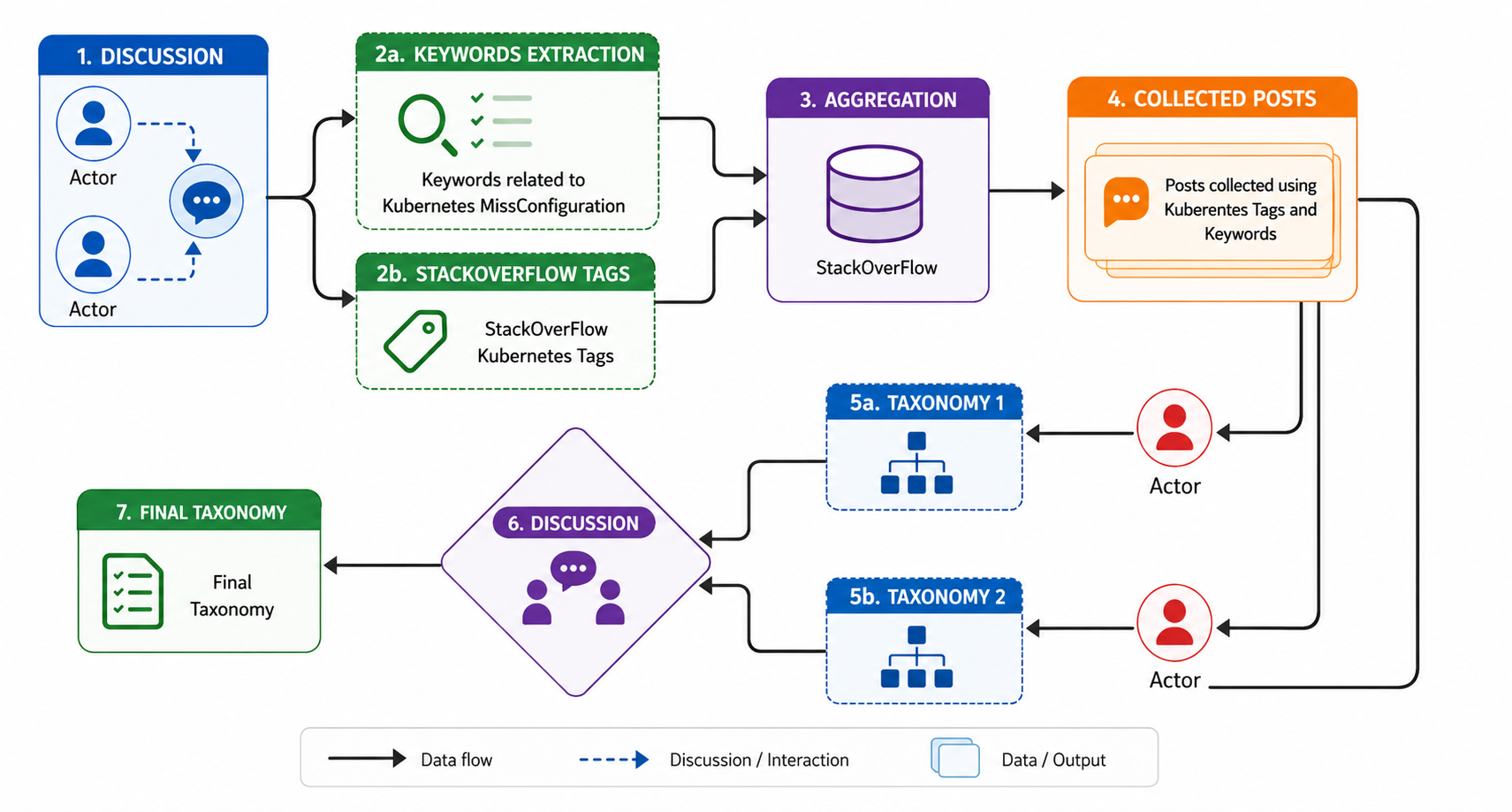} 
\caption{Kubernetes Taxonomy Development Process}
\label{fig:FigApp01}
\end{figure}

To address the first research question, we collected Kubernetes configuration-related posts from Stack Overflow, as shown in Figure~\ref{fig:FigApp01}. Stack Overflow was chosen for its wide range of technical discussions and its widespread use by software developers. The platform attracts a broad spectrum of practitioners, from beginners to experts, offering a diverse and realistic view of the challenges encountered in practice. This diversity enables the identification of configuration errors arising at varying levels of expertise.

To construct a targeted dataset, we extracted all available tags from Stack Overflow and manually reviewed them to identify those specifically related to Kubernetes while minimizing overlap with unrelated topics. The final set of selected tags included:
\textbf{kubernetes, k8s, docker, kube, kubectl and nginx-ingress}.

In order to improve the data set and focus on configuration-related issues, we developed a list of relevant keywords through an expert-driven process. Two domain specialists proposed an initial set of keywords based on their knowledge of Kubernetes security vulnerabilities and misconfigurations. These keywords were refined through iterative discussions to reach a consensus on the final selection:

\textbf{[Misconfiguration, Privilege Escalation, Network Exposure, Access Control, Secret Management, Cluster Access, RBAC, Insecure Network, Ingress Controller, Namespace Isolation, Dangerous Permissions, Escalation Control, Root Filesystem, Privileged Mode, Default Namespace, Root Access, Memory Limits]}.

These keywords were systematically applied throughout the data collection process. Using the selected tags and validated keywords, we retrieved Stack Overflow posts addressing Kubernetes misconfigurations. Initially, 4,785 posts were collected based on Kubernetes-related tags. To improve relevance, we filtered the data set to include only posts containing at least one validated keyword, resulting in a final dataset of 681 posts. These posts, spanning from 2015 to 2024, specifically focus on Kubernetes configuration issues.

\begin{figure}[h!]
\centering
\includegraphics[width=9cm, height=6cm]{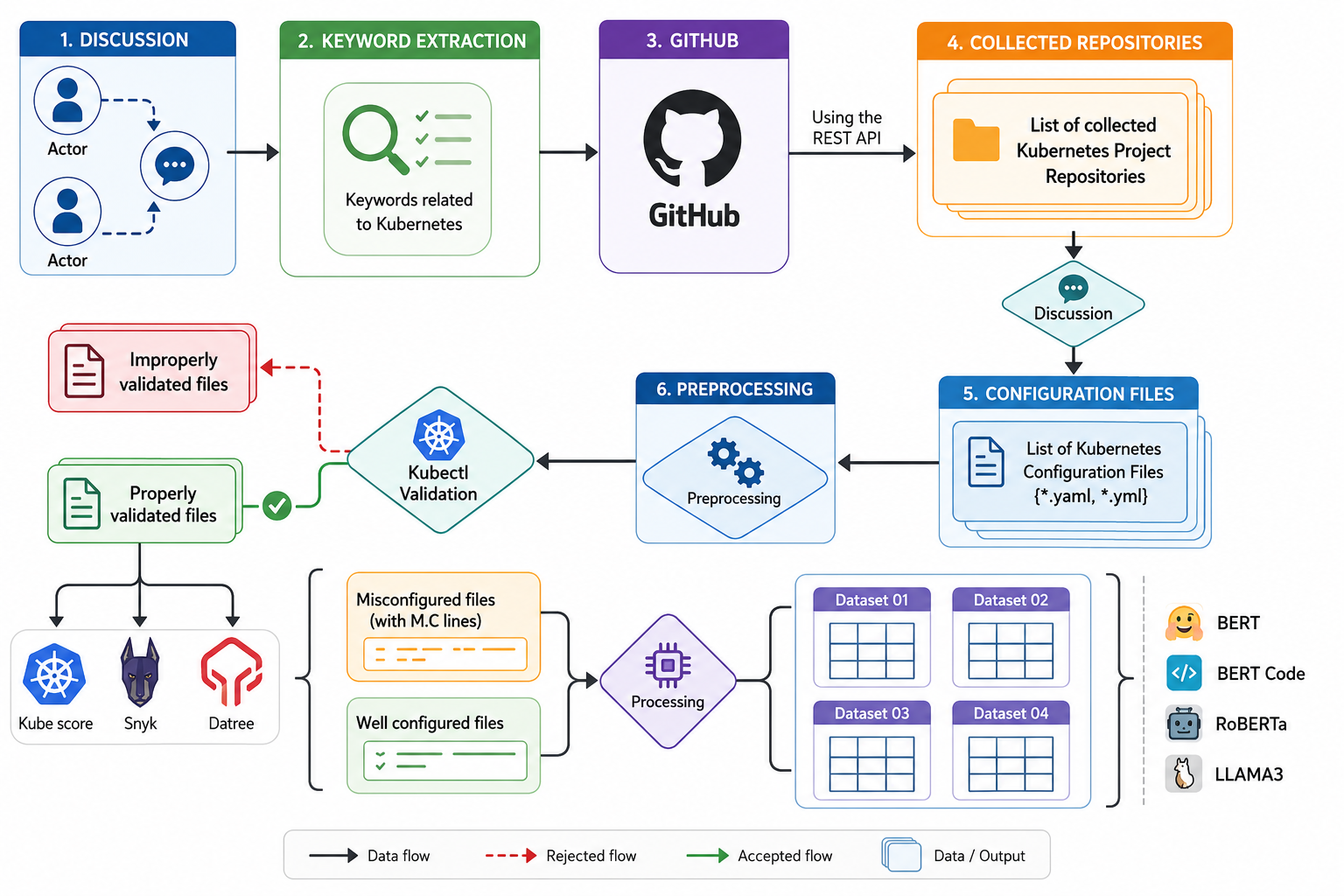} 
\caption{Kubernetes Misconfiguration Detection Flow}
\label{fig:FigApp02}
\end{figure}

To address the second research question, we collected the Kubernetes configuration file from GitHub using the procedure indicated in Figure \ref{fig:FigApp02}. Using the previously selected set of keywords, we identified repositories related to Kubernetes and retrieved all files with the extensions .yaml and .yml. To automate this process, we integrated the GitHub REST API into our workflow. After extracting the configuration files, we performed an initial validation step to ensure their syntactic correctness. All files were processed using this command:

{\small
\begin{verbatim}
kubectl apply --dry-run=client -f <file.yaml>
\end{verbatim}
}

Only files that passed this validation step were included in the dataset, ensuring a high level of data integrity.

Following validation, we refined the dataset by removing comments, unnecessary lines, and other irrelevant content. For the labeling process, we employed three Kubernetes misconfiguration detection tools: Snyk \cite{b9}, Kube Score \cite{b8}, and Datree \cite{b10}. These tools were selected based on their ability to identify the highest number of misconfigurations among the options evaluated. Specifically, Kube Score provides a comprehensive assessment of configuration practices in accordance with official Kubernetes guidelines. Snyk specializes in detecting security vulnerabilities, particularly those related to container image dependencies. Datree focuses on enforcing customizable policies to ensure compliance with best practices. Our labeling strategy adopts a conservative approach, any issue flagged by at least one tool is treated as a misconfiguration. By integrating these tools, we captured a broad spectrum of misconfigurations, including security weaknesses, suboptimal configurations, and deviations from Kubernetes standards. After this process, the initial labeled dataset was used to generate four distinct datasets :

\begin{itemize}

    \item \textbf{Dataset 1 (13723 entries):} This dataset contains all collected configuration files. A file is considered misconfigured if at least one detection tool flags it. If no tool finds any misconfiguration, the file is considered safe.
    
    \item \textbf{Dataset 2 (26984 entries):} In this dataset, lines classified as misconfigured by any of the three tools are designated as misconfigured. To balance the dataset, an equal number of lines from the same object type, which are not flagged by any of the tools, are designated as safe.
    
    \item \textbf{Dataset 3 (32956 entries):} This dataset is similar to Dataset 2 but excludes the four lines following each detected misconfiguration, as these lines are considered potentially suspect. The safe portions of the dataset are defined as sections that are neither flagged as misconfigured nor located within these suspect regions.
    
    \item \textbf{Dataset 4 (276520 entries):} In this dataset, configuration files are organized by object type.  Objects identified as misconfigured are labeled accordingly. We observed that the number of safe objects significantly exceeds the number of misconfigured objects. To mitigate this class imbalance, we developed a tool to inject additional misconfigurations, thereby improving the representativeness of the dataset.
\end{itemize}

To mitigate the class imbalance in Dataset 4, we applied three targeted mutation operations: deletion, insertion, and modification. These operations simulate realistic misconfigurations by removing security-relevant fields (such as securityContext), inserting risky parameters (such as privileged: true), or altering safe values (for example, changing runAsNonRoot to false).

Each mutated object was validated for syntax correctness and rescanned using the same three detection tools employed during the initial labeling. Only configurations flagged as faulty by at least one tool were retained.

Due to space constraints, we do not provide a detailed description of the synthetic generator. However, we emphasize that after each misconfiguration was injected, a syntax validation step was performed, followed by a rescan using labeling tools. Only configurations flagged by at least one of these tools were retained.

\subsection{Modeling}
In this study, we used state-of-the-art models to improve the detection of misconfigurations in Kubernetes environments. These models were selected based on their proven effectiveness in natural language processing, code understanding, and anomaly detection, as demonstrated in previous research. Specifically, we focused on BERT, CodeBERT, RoBERTa, and LLaMA3, each offering unique advantages. The following sections detail their key characteristics, benefits, and the hyperparameter settings used in our experiments.

\subsubsection{\textbf{BERT (Bidirectional Encoder Representations from Transformers)}}

BERT was selected for its bidirectional context modeling, which has proven highly effective in detecting anomalies in various studies , including \cite{b43}, \cite{b44}, \cite{b45}, \cite{b46}, \cite{b47}. Its ability to process context in both directions enables a deeper understanding of token relationships, making it particularly suitable for identifying misconfigurations in Kubernetes configuration files where subtle contextual dependencies play a critical role. For this study, we utilized the pre-trained bert-base-uncased model, fine-tuned specifically to capture nuanced patterns within Kubernetes configurations.

\subsubsection{\textbf{CodeBERT}}
CodeBERT was selected for its proficiency in processing tasks that encompass both natural language and source code, rendering it particularly effective in identifying misconfigurations within YAML files. As an extension of BERT, CodeBERT has been pre-trained on a diverse corpus comprising both source code and natural language, thereby augmenting BERT’s capability to comprehend code syntax and semantics. Furthermore, empirical studies, including \cite{b107} and \cite{b108}, have demonstrated its strong performance in the detection of bugs and anomalies.

\subsubsection{\textbf{RoBERTa (Robustly optimized BERT approach)}}

RoBERTa was selected due to its advanced training methodology, which involves pretraining on a substantially larger corpus (160 GB of text \cite{b100}, compared to BERT’s 20 GB \cite{b101}) and the elimination of the Next Sentence Prediction (NSP) objective. These modifications enhance RoBERTa’s ability to capture deeper contextual dependencies, enabling it to identify complex misconfigurations that span multiple lines or involve intricate interdependencies with greater precision.

\subsubsection{\textbf{LlaMA3}}
LLaMA3 was selected for its ability to process long-range dependencies and structured data, making it effective in detecting Kubernetes misconfigurations. Pretrained on a large corpus, it captures semantic nuances in YAML and JSON configurations. Several studies \cite{b48}, \cite{b49}, \cite{b50} have demonstrated its high performance in vulnerability detection, especially in identifying misconfigurations that involve multiple lines or complex dependencies between parameters.

\subsection{Fine-Tuning}
All models were fine-tuned for the task of misconfiguration detection using a consistent hyperparameter setup: a learning rate of 1e-4, a batch size of 8, and a weight decay of 0.01 over five epochs. To enhance efficiency and performance, quantization and LoRA (Low-Rank Adaptation) techniques were applied. Quantization was performed using 4-bit configurations via the \texttt{BitsAndBytes} library, employing the \texttt{nf4} quantization type optimized for normally distributed weights. Additional optimizations, including double quantization and \texttt{torch.bfloat16} computations, further reduced memory requirements without compromising performance. LoRA was implemented with a low-rank dimension of 160 and an alpha scaling factor of 8, targeting attention layers such as query, key, and value projections. A dropout rate of 0.05 was introduced for regularization, improving the models' adaptability to the structured and contextual nature of Kubernetes configurations while maintaining computational efficiency. This combined approach enabled the models to achieve high accuracy in detecting Kubernetes misconfigurations while being computationally efficient and suitable for real-world deployment \cite{lora1} \cite{lora2}.

For BERT, CodeBERT, and RoBERTa, a supervised learning approach was adopted, where each model was trained on labeled datasets to detect misconfigurations in Kubernetes configurations. This process involved learning to map specific features of the configuration files to corresponding labels, such as misconfigured or well-configured. The supervised training allowed the models to leverage contextual relationships within the configuration files and make accurate predictions based on a corpus of annotated examples. LLaMA3, on the other hand, was fine-tuned on a similar labeled dataset of Kubernetes configuration files, where configuration codes served as input prompts with labels indicating their configuration state. This enabled the model to learn patterns and anomalies in the data, facilitating its ability to detect misconfigurations. The training process involved preprocessing configuration files, tokenizing them using the LLaMA3 tokenizer, and addressing imbalanced label distributions with weighted loss functions. To optimize efficiency, 4-bit quantization and LoRA techniques were employed, reducing memory requirements while maintaining model performance.

During inference, all models analyzed configuration files as input prompts, generating predictions by evaluating logits outputs. These predictions were compared against true labels, and metrics such as balanced accuracy and classification reports were used to evaluate their performance. The contextual understanding of the models allowed them to identify both explicit syntax errors and more subtle logical inconsistencies in Kubernetes configurations.  In the following sections, we evaluate the performance of these models across different datasets, providing a detailed analysis of their strengths and weaknesses in relation to our research objectives.

\section{Evaluation}
In this section, we detail our experimental methodology, present the obtained results, and conduct a comprehensive evaluation of state-of-the-art tools. We begin by articulating the key research questions that define the scope of our investigation. Through a rigorous analysis of the results, we assess the effectiveness of our proposed approach and its potential contributions to advancing the field.

\subsection{Research Questions}

To structure our evaluation, we define the following research questions, which serve as the foundation for our experimental design and analysis:

\begin{itemize}
    \item \textbf{RQ1 :}  What key categories should be included in a comprehensive taxonomy of Kubernetes misconfigurations, considering the common concerns and inquiries raised by developers?
    
     \item \textbf{RQ2 :} What is the frequency of Kubernetes misconfigurations across object types, how can their risks be evaluated, and how effectively do LLMs detect them?
     
    \item \textbf{RQ3 :} How do existing misconfiguration detection tools compare, and how does our approach perform relative to state-of-the-art methods?
   
\end{itemize}

\subsubsection{\textbf{Evaluation and results for RQ1}}
To address this research question, we adopted a systematic approach that involved collecting and analyzing relevant data from Stack Overflow posts published between 2015 and 2024, as presented in Section ~\ref{sec:DataCollection}.

Subsequently, two participants independently analyzed each post to classify them based on the identified misconfiguration patterns. When classification discrepancies arose, multiple structured discussion sessions were held to reach a consensus. These sessions were not limited to resolving disagreements at the post level but also focused on refining the categorization schema itself. Specifically, the participants reviewed and unified the naming of categories to ensure consistency, merged overlapping categories when they referred to similar types of misconfiguration, and split broader categories into more specific categories when necessary to improve granularity and clarity. These decisions were guided by a set of internal criteria, including semantic similarity, contextual overlap, the object affected by the misconfiguration, the associated risk type, and the clarity of the distinction between categories. As a result of this iterative process, the resulting taxonomy is inherently adaptable. New misconfiguration cases can be incorporated into existing categories when relevant, or the taxonomy can be extended by introducing additional categories to reflect emerging issues. This flexibility ensures that the classification remains applicable despite the continuous evolution of Kubernetes APIs, plugins, and deployment environments.

Through this iterative process, documented and illustrated in Figure~\ref{fig:FigApp01}, the final taxonomy was refined and agreed upon. The outcome of this classification is summarized in Table~\ref{tab:kubernetes-Manual-categories}, while Figure~\ref{fig:FigApp01} provides a comprehensive overview of the categorization and decision-making process.

\begin{table}[ht]
    \centering
    \begin{tabular}{|p{0.25\linewidth}|p{0.65\linewidth}|} 
        \hline
        \textbf{Category} & \textbf{Description} \\
        \hline
        \textbf{Container Security} & Measures to restrict access, prevent privilege escalation, and secure container capabilities and privileges. \\
        \hline
        \textbf{Resource Management} & Configurations to optimize resource utilization such as CPU and memory, and ensure appropriate limits. \\
        \hline
        \textbf{Probe Configuration} & Ensuring each container has configured liveness and readiness probes to maintain application availability and stability. \\
        \hline
        \textbf{Permissions and Access} & Control permissions and access, including management of service account tokens and limiting container capabilities. \\
        \hline
        \textbf{Image Configuration} & Ensuring each container image uses a fixed (tagged) version for reproducibility and update management. \\
        \hline
        \textbf{API Usage} & Avoiding the use of deprecated APIs that may pose security or stability risks. \\
        \hline
        \textbf{Network and Cryptography Configuration} & Configurations to ensure encryption, enforcement of HTTPS, and other network security and cryptography measures. \\
        \hline
    \end{tabular}
    \caption{Results of Manual Process: Categories of Kubernetes Configurations}
    \label{tab:kubernetes-Manual-categories}
\end{table}

\vspace{-1em}
\textbf{Container Security}
This category covers important security measures necessary to protect container environments. It includes settings limit access, prevent unauthorized privilege escalation, and control container permissions. A key focus is ensuring that containers do not run with root privileges by setting security contexts and applying Pod Security Policies (PSPs) to enforce security rules. For example, Post \footnote{https://stackoverflow.com/questions/75375548}, from a Stack Overflow discussion, highlights the risks of running containers in privileged mode, which is generally discouraged because it grants excessive permissions.

\textbf{Resource Management}
Proper resource management is crucial for maintaining Kubernetes cluster performance and stability. This includes optimizing CPU and memory usage, setting resource limits, and preventing contention. Effective management ensures containers receive adequate resources without overutilization. \footnote{https://stackoverflow.com/questions/32034827} highlights challenges in enforcing resource limits, particularly when usage exceeds predefined limits after pod creation.

\textbf{Probe Configuration}
This category focuses on configuring liveness and readiness probes to maintain application health and stability. Properly set probes ensure accurate monitoring of container health and readiness to handle traffic, preventing disruptions caused by misconfigurations. For example, this Stack Overflow post\footnote{https://stackoverflow.com/questions/78690753} highlights a discussion on implementing Kubernetes liveness probes in applications relying on external databases, evaluating whether to include database connectivity checks or use independent health checks.

\textbf{Permissions and Access}
Managing permissions and access controls is crucial for securing Kubernetes environments. This includes configuring Role-Based Access Control (RBAC) roles, service account tokens, and other access controls to prevent overly permissive roles and security breaches. For example \footnote{https://stackoverflow.com/questions/55620567},an RBAC error preventing an ingress controller from obtaining the necessary permissions, indicating that the cluster’s authorization settings are blocking access.

\textbf{Image Configuration}
Proper container image management is crucial for reproducibility and security in Kubernetes. Relying on the mutable latest tag can lead to unpredictable deployments, inconsistency, and security risks. Unscanned or outdated images can expose the system to vulnerabilities, making regular updates and vulnerability scanning essential. For example, in this post\footnote{https://stackoverflow.com/questions/76587402}, a user encounters an error when using the latest tag in a Kubernetes Job, highlighting the risks associated with mutable tags across different deployment scenarios.

\textbf{API Usage}
Secure access to the Kubernetes API is crucial for managing cluster resources in accordance with security best practices. Proper configuration of authentication and authorization mechanisms helps prevent unauthorized access and ensures safe interaction with the cluster. The post\footnote{\url{https://stackoverflow.com/questions/32294007}} highlights challenges encountered when accessing the Kubernetes API server securely via \texttt{kubectl} on macOS.

\textbf{Network and Cryptography Configuration}
Effective network and encryption configurations are essential for securing communications within Kubernetes clusters. This includes enforcing HTTPS, encrypting sensitive data, and applying cryptographic measures to protect data in transit. The post \footnote{\url{https://stackoverflow.com/questions/60195653}} highlights an issue where a health check pod encounters an SSL handshake error when connecting to a LoadBalancer service.

These categories offer a comprehensive perspective on the diverse categories of Kubernetes misconfigurations, highlighting the extensive scope of potential issues and underscoring the critical role of rigorous configuration management. Effectively addressing these aspects can substantially enhance the security and resilience of Kubernetes environments.

The framework is also designed to support future extensibility. Newly discovered misconfigurations can be assigned to existing categories where appropriate, or new categories can be introduced to accommodate emerging issues. This flexibility ensures that the taxonomy remains applicable over time, even as Kubernetes evolves through changes in API versions, third-party integrations, and deployment practices.

\subsubsection{\textbf{Evaluation and Results for RQ2}}

Before addressing this research question and evaluating the models' performance in detecting Kubernetes misconfigurations, we first aimed to identify the components of the Kubernetes ecosystem most susceptible to misconfigurations and those posing the greatest security and operational risks. To accomplish this, we conducted an analysis focusing on two key aspects:

\paragraph{\textbf{Frequency of Misconfigurations by object}}
To investigate this aspect, we conducted an analysis using three widely recognized Kubernetes configuration scanning tools: Kubescore, Snyk, and Datree. These tools employ rule-based detection mechanisms to identify potential misconfigurations across various Kubernetes objects. The selection of these tools was based on their methodological diversity, effectiveness in detecting a broad spectrum of misconfigurations, and established credibility within the industry. Using multiple tools, our goal was to obtain a more comprehensive and unbiased assessment of configuration issues.

To ensure an accurate estimation of unique misconfigurations per object and to eliminate redundancy caused by overlapping detections, we manually reviewed the results produced by all three tools. Misconfigurations flagged by multiple tools for the same object, of the same type, and at the same location were counted only once. Given the need for detailed inspection and precise validation, this analysis was conducted on a smaller, high-quality dataset to ensure feasibility.

Specifically, for addressing the first part of RQ2, which concerns the number of unique misconfigurations per object, we used the official Helm charts repository for Kubernetes\footnote{\url{https://github.com/helm/charts/tree/master}}. The first four datasets described earlier were excluded from this phase due to their large size, which made manual de-duplication across tens of thousands of configuration files impractical.

Helm charts are widely adopted in the Kubernetes ecosystem and offer standardized templates for deploying complex cloud-native applications. This made them an appropriate and representative source for our analysis, providing both diversity in Kubernetes object types and relevance to real-world deployment practices.

The final collection included 2,097 Kubernetes configuration files, which were systematically analyzed using all three tools. The automated scans yielded the following results: Kubescore identified 1,082 misconfigurations, Snyk detected 1,058, and Datree reported 611. To ensure the reliability of our findings, we performed a manual review to eliminate redundant detections, considering each misconfiguration flagged by multiple tools as a single instance, in order to determine the total number of unique misconfigurations per object. Furthermore, we conducted an in-depth examination of all misconfigured files to classify affected Kubernetes objects systematically. A detailed distribution of misconfigurations by object type is presented in Table \ref{table:misconfigurations-by-components}, where the lines are sorted by the total frequency of misconfigurations per object.

\begin{table}[h!]
\centering
\begin{tabular}{|l|c|c|c|c|}
\hline
\textbf{Object} & \textbf{Kube-score} & \textbf{Snyk} & \textbf{Datree} & \textbf{total} \\
\hline
Deployment          & 593 & 482 & 234 & 779 \\
Pod                 & 211 & 249 & 142 & 245 \\
NetworkPolicy       & 216 & 96  & 107 & 197 \\
Service             & 63  & 45  & 28  & 105 \\
Role/ClusterRole    & 57  & 48  & 36  & 93  \\
StatefulSet         & 24  & 32  & 19  & 57  \\
ConfigMap           & 2   & 11  & 3   & 12  \\
Secret              & 0   & 11  & 5   & 11  \\
Others              & 81  & 93  & 37  & 187 \\
\hline
\textbf{Total}      & \textbf{1082} & \textbf{1058} & \textbf{611} & \textbf{1686} \\
\hline
\end{tabular}
\caption{Misconfiguration Detection by Kubernetes objects}
\label{table:misconfigurations-by-components}
\end{table}

The results presented in Table \ref{table:misconfigurations-by-components} presents a quantitative analysis of misconfigurations identified by Kube-score, Snyk, and Datree across various Kubernetes object types. The results show that Kube-score consistently detected the highest number of misconfigurations, particularly in Pods, NetworkPolicies, and Deployments, surpassing both Snyk and Datree. Notably, Kube-score identified 593 misconfigurations in Deployments, compared to 482 by Snyk and 234 by Datree.

Furthermore, Pods and NetworkPolicies exhibited a significant number of misconfigurations, emphasizing key areas where configuration management practices need improvement. These findings suggest that different tools have varying detection capabilities, highlighting the importance of using multiple tools to comprehensively assess Kubernetes security and configuration integrity.

Overall, the results highlight that each tool has different rule coverage and detection capabilities, leading to varying numbers of detected misconfigurations per component. The total column, which accounts for unique misconfigurations, is higher than individual tool detections, emphasizing the complementary nature of these tools. This suggests that relying on a single tool may not be sufficient for comprehensive security and configuration analysis, and using multiple tools can provide better coverage.

\paragraph{\textbf{Evaluation of misconfiguration risks by object}}

To address this aspect of our research question, which aims to identify the Kubernetes ecosystem objects most susceptible to high-risk misconfigurations, we conducted an analysis using three selected misconfiguration detection tools. This assessment focused specifically on misconfigurations classified as moderate, high, or very high risk. We scanned a dataset of 2,097 configuration files, and based on the results, we categorized an object as susceptible to a specific severity level if more than half of the detected misconfigurations within it fell into that severity category. Otherwise, the object was considered susceptible to the two most frequently detected severity levels within its type.

It is important to clarify that the severity (risk) level of each misconfiguration is based on the classification provided by the detection tools referenced in our study. These tools determine severity levels using predefined rules that incorporate expert knowledge, industry best practices, and references to known security standards such as reported CVEs and CWEs. We did not modify or reinterpret these classifications. Instead, we relied directly on the severity levels reported by the tools to infer the overall risk profile associated with each Kubernetes object type.

In addition to severity, we also associated each Kubernetes object with the most recurrent category of misconfiguration observed during the analysis. The categories such as Permissions and Access, Resource Management, Probe Configuration, and Network and Cryptography Configuration offer a qualitative dimension to the risk landscape, helping to identify the dominant misconfiguration pattern specific to each object type. The detailed results are summarized in Table \ref{table:risk}.

The analysis of Kubernetes misconfigurations reveals that certain objects exhibit a higher propensity for severe security risks, necessitating strict configuration management. ClusterRoles, Services, and RBAC are the most vulnerable, with misconfigurations leading to critical security threats. ClusterRoles, frequently misconfigured, pose risks of unauthorized access, while Services and RBAC errors can expose applications to external threats and compromise overall security. StatefulSets and DaemonSets present moderate to high risks, affecting data integrity and node stability. Roles and Deployments, though less misconfigured, can still lead to unauthorized access and service disruptions. Other objects, like ConfigMaps and Secrets, have variable risk levels, underscoring the need for comprehensive security measures across all Kubernetes components.

In conclusion, while multiple Kubernetes objects are susceptible to misconfigurations, ClusterRoles, Services, and RBAC configurations represent the most critical concerns. Addressing these vulnerabilities through rigorous configuration management is essential to ensure the security, stability, and resilience of Kubernetes environments.
Furthermore, the identification of dominant misconfiguration categories per object can guide the development of targeted mitigation strategies and more effective policy enforcement.

\begin{table*}[ht]
\centering
\begin{tabular}{|p{1.5cm}|p{11cm}|p{3cm}|p{2cm}|}
\hline
\textbf{K8S Object} & \textbf{Description} & \textbf{Common Misconfiguration Category} & \textbf{Risk Level} \\ \hline
\textbf{ClusterRole} & Defines permissions across the entire cluster. Misconfigurations can lead to unauthorized access to sensitive objects, compromising cluster security. & Permissions and Access & Very High \\ \hline
\textbf{StatefulSet} & Manages stateful applications. Misconfigurations can impact data availability and integrity. & Resource Management & Moderate to High \\ \hline
\textbf{DaemonSet} & Ensures a pod runs on each node. Misconfigurations can cause cluster overload or instability. & Resource Management & Moderate to High \\ \hline
\textbf{Role} & Defines permissions within a specific namespace. Misconfigurations can allow unauthorized object manipulation. & Permissions and Access & Moderate \\ \hline
\textbf{Deployment} & Manages stateless application deployments. Misconfigurations can result in service interruptions. & Probe Configuration & Moderate \\ \hline
\textbf{Service} & Exposes applications to networks. Misconfigurations can expose services to external attacks. & Network and Cryptography Configuration & High \\ \hline
\textbf{RBAC} & Manages user and group permissions. Misconfigurations can grant unauthorized access, affecting cluster security. & Permissions and Access & Very High \\ \hline
\textbf{Others} & Various configurations like ConfigMaps and Secrets. Risk varies based on object usage. & Image Configuration & Moderate to High \\ \hline
\end{tabular}
\caption{Analysis of Kubernetes Object Misconfigurations, Common Categories, and Associated Risks}
\label{table:risk}
\end{table*}

Following this analysis, we evaluatedd the performance of multiple models across four datasets measuring accuracy, precision, and recall. Table \ref{table:evalmodels} summarizes the results, providing insights into model behavior under varying data distributions. Accuracy measures overall correctness, precision reflects the reliability of positive predictions, and recall captures the ability to identify relevant cases. These metrics offer a comprehensive assessment of each model’s effectiveness in detecting Kubernetes misconfigurations.

\begin{table}[h]
\centering
\begin{tabular}{| p{1.3cm} | p{1.3cm} | p{0.8cm} |p{0.8cm} |p{0.8cm} |p{0.8cm} |p{0.5cm} |p{0.5cm} |}

\hline
\centering \textbf{Model} & \centering Metrics & \centering \textbf{D1} & \centering \textbf{D2} & \centering \textbf{D3} &  \textbf{D4} \\
\hline
BERT & \begin{tabular}{c}
Precision \\
Recall \\
F1 score \\
\end{tabular} & 
\begin{tabular}{c}
 12.94 \\
 49.69 \\
 20.53 \\
\end{tabular} & 
\begin{tabular}{c}
 91.42 \\
 92.43 \\
 91.92 \\
\end{tabular} & 
\begin{tabular}{c}
 87.96 \\
 92.59 \\
 90.22\\
\end{tabular} & 
\begin{tabular}{c}
 99.74 \\
 99.94\\
 99.62 \\
\end{tabular} \\
\hline
CodeBERT &  \begin{tabular}{c}
Precision \\
Recall \\
F1 score \\
\end{tabular} & 
\begin{tabular}{c}
 98.85 \\
 50.01 \\
 66.42 \\
\end{tabular} & 
\begin{tabular}{c}
 89.76 \\
 94.34 \\
 92.00 \\
\end{tabular} & 
\begin{tabular}{c}
 89.15\\
 90.98\\
 90.06 \\
\end{tabular} & 
\begin{tabular}{c}
 99.73 \\
 99.93 \\
 99.83 \\
\end{tabular} \\
\hline
RoBERTa &  \begin{tabular}{c}
Precision \\
Recall \\
F1 score \\
\end{tabular} & 
\begin{tabular}{c}
 98.84 \\
 50.01 \\
 66.42 \\
\end{tabular} & 
\begin{tabular}{c}
 92.32 \\
 89.85 \\
 91.07 \\
\end{tabular} & 
\begin{tabular}{c}
 86.32 \\
 91.42 \\
 88.79 \\
\end{tabular} & 
\begin{tabular}{c}
 99.75 \\
 99.93 \\
 99.84 \\
\end{tabular} \\
\hline
LlaMA3 &  \begin{tabular}{c}
Precision \\
Recall \\
F1 score \\
\end{tabular} & 
\begin{tabular}{c}
 98.63 \\
 50.78 \\
 67.04 \\
\end{tabular} & 
\begin{tabular}{c}
 93.53 \\
 94.34 \\
 93.93 \\
\end{tabular} & 
\begin{tabular}{c}
 93.54 \\
 96.15 \\
 94.83 \\
\end{tabular} & 
\begin{tabular}{c}
 99.84 \\
 99.91 \\
 99.88 \\
\end{tabular} \\
\hline
\end{tabular}
\caption{Performance Metrics Summary for Each Model Across Datasets}
\label{table:evalmodels}
\end{table}

The results presented in Table 4 reveal varying levels of performance across different models when applied to various datasets, each characterized by distinct preprocessing and data-handling strategies.

In Dataset 1, which underwent minimal preprocessing, both BERT and LLaMA3 exhibited relatively low precision and F1 scores. This outcome suggests that these models struggled to accurately identify misconfigurations in a dataset with a complex or less structured format. In contrast, CodeBERT and RoBERTa demonstrated marginally higher precision but lower recall, indicating their ability to detect certain misconfigurations while potentially overlooking others.

Datasets 2 and 3 incorporated more extensive preprocessing. Dataset 2 included explicit annotations for misconfigured and well-configured lines, leading to a significant performance improvement across all models, with F1 scores approaching 90\%. These results suggest that the models were more effective when handling structured data with clearly defined labels. In Dataset 3, an additional preprocessing step was introduced by incorporating four subsequent lines following each misconfiguration, based on the assumption that these lines could be influenced by the misconfiguration. This approach further enhanced model performance, as reflected in consistently high F1 scores, indicating improved capability in processing extended contextual information.

For Dataset 4, segmentation was performed based on object types, ensuring a balanced distribution of misconfigured and well-configured instances. This rigorous segmentation resulted in near-perfect precision and recall across all models, with LLaMA3 demonstrating particularly strong performance. These findings indicate that when data is well-structured and balanced, all models especially LLaMA3 excel in accurately detecting misconfigurations.

Overall, while performance varied across datasets, LLaMA3 and RoBERTa consistently emerged as the most effective models. Notably, in Dataset 4, where the dataset was optimally segmented and balanced, both models achieved near-optimal results in terms of precision and recall.

\paragraph{\textbf{Granular Failure Analysis by Misconfiguration Type}}

To further assess the reliability of our best-performing models using the best segmentation method, we conducted a fine-grained error analysis of the LLaMA3 model on Dataset 4, focusing on the types of misconfigurations that were either frequently misclassified or consistently detected. This complementary evaluation aimed to uncover which Kubernetes misconfiguration patterns are most challenging for the model and which are effectively captured.

By correlating the false positives and false negatives with the underlying Kubernetes objects and their associated misconfiguration types, as defined in our proposed taxonomy, we identified clear trends in model performance across misconfiguration scenarios. The analysis revealed that LLaMA3 exhibited high accuracy in detecting misconfigurations related to \texttt{Probe Configuration} and \texttt{Resource Management}, particularly within \texttt{Deployment} and \texttt{StatefulSet} objects. These cases were typically well-structured and followed consistent patterns in configuration syntax, enabling the model to generalize effectively across similar instances.

In contrast, misconfigurations involving access control components particularly those found in \texttt{Role}, \texttt{ClusterRole}, and \texttt{RBAC} configurations proved to be more challenging. These errors often involved subtle permission settings or non-obvious access scopes, which led to a higher rate of false negatives. Similarly, \texttt{NetworkPolicy} misconfigurations also contributed to reduced precision, likely due to their diverse and context-dependent specifications, which can be semantically valid yet operationally insecure.

Moreover, objects categorized under \texttt{Others}, such as \texttt{ConfigMap} and \texttt{Secret}, showed inconsistent detection performance. Their configuration syntax tends to be lightweight or unstructured, and the security impact of their misconfiguration often depends on broader deployment context factors that may not be fully captured by the model’s.

Overall, the findings highlight that while LLaMA3 performs exceptionally well on syntactically clear and structurally consistent configuration issues, its performance diminishes for nuanced or semantically complex misconfigurations, particularly those involving access control and network behavior. These insights can guide the development of complementary rule-based validation layers or fine-tuning strategies to mitigate model weaknesses in future deployments.

\subsubsection{\textbf{Evaluation and results for RQ3}}

To rigorously assess the effectiveness of our approach, we conducted a comprehensive comparative analysis with several widely used rule-based misconfiguration detection tools. For this evaluation, we employed a standardized benchmark dataset derived from the official Helm Charts repository for Kubernetes. This repository was explicitly excluded from our model's training data to ensure an unbiased evaluation and prevent data leakage.

We selected six prominent tools representative of the current state-of-the-art in static analysis for Kubernetes configurations: Snyk, Kube Score, Datree, Checkov, KubeLinter \cite{b104}, and Terrascan \cite{b106}. Each tool was applied to the benchmark dataset, and their respective detection capabilities were evaluated based on the total number of misconfigurations identified.

Following this, we applied our best performing LLMs to the same dataset, after fragmenting the configuration files into a subset of objects using the same method employed in the creation of the fourth dataset.  This ensured a fair and consistent comparison across all tools. This two-phase evaluation highlighted both the limitations of rule-based tools and the practical advantages of LLM-based analysis.

The results show that our model significantly outperforms existing tools by covering a wider range of misconfigurations and handling more complex cases that rule-based tools do not detect. As detailed in Table \ref{tab:component_comparison}, our model identified 3,209 misconfigurations, compared to 1,058 by Snyk and 1,082 by Kube Score. This substantial improvement is primarily attributed to the LLM's ability to identify novel and complex misconfiguration patterns that are often overlooked by tools relying on static, rule-based logic.

In particular, we did not include a direct comparison with GenKubeSec, as neither its dataset nor its detection model is publicly accessible. However, based on the performance metrics reported in their publication, our approach offers broader coverage and superior detection capabilities, particularly in analyzing diverse types of Kubernetes configuration files. In contrast, GenKubeSec remains limited in its applicability to certain configuration formats.

To enhance the comparative evaluation, we conducted an in-depth analysis of the misconfigurations identified by each tool, categorizing them according to the type of affected Kubernetes object. This classification offers valuable insight into the object-specific detection capabilities of each approach and further highlights the superior performance of our model. Table \ref{tab:component_comparison} summarizes the number of misconfigurations detected per Kubernetes object type, as well as the overall total.

\begin{table}[ht]
\centering

\begin{tabular}{| p{2cm} | p{0.5cm} | p{0.5cm} |p{0.5cm} |p{0.5cm} |p{0.5cm} |p{0.5cm} |p{0.5cm} |}
\hline
\centering \textbf{objects} & \rotatebox{90}{\textbf{Snyk}} & \rotatebox{90}{\textbf{Kube Score}} & \rotatebox{90}{\textbf{Datree}} & \rotatebox{90}{\textbf{Checkov}} & \rotatebox{90}{\textbf{KubeLinter}} & \rotatebox{90}{\textbf{Terrascan}} & \rotatebox{90}{\textbf{Best Model}} \\
\hline
Pod                & 249    & 206   & 142    & 94    & 135    & 79    & 713   \\
NetworkPolicy      & 96     & 127   & 107    & 63    & 51     & 41    & 466   \\
Role/ClusterRole   & 48     & 54    & 36     & 9     & 13     & 25    & 137   \\
Service            & 45     & 62    & 28     & 17    & 9      & 18    & 102   \\
Deployment         & 473    & 543   & 234    & 182   & 62     & 106   & 1436  \\
StatefulSet        & 32     & 24    & 19     & 13    & 4      & 17    & 78    \\
ConfigMap          & 11     & 2     & 3      & 3     & 2      & 1     & 19    \\
Secret             & 11     & 0     & 5      & 1     & 4      & 6     & 17    \\
Others             & 93     & 64    & 37     & 27    & 18     & 23    & 241   \\
Total              & 1058   & 1082  & 611    & 409   & 298    & 316   & 3209   \\
\hline
\end{tabular}
\caption{Comparison of Misconfiguration Detection Across Kubernetes objects and Tools}
\label{tab:component_comparison}
\end{table}

Table \ref{tab:component_comparison} shows that different tools vary in how well they detect misconfigurations in Kubernetes objects. Traditional tools like Snyk, Kube Score, and Datree find the most misconfigurations in Deployments and Pods. This is likely because these important objects have more rules to check.

On the other hand, tools such as KubeLinter \cite{b104} and Terrascan \cite{b106} find fewer misconfigurations, especially in simpler objects like ConfigMaps and Secrets. This suggests these tools are less focused on these resource types.

Our best model detects many more misconfigurations than other tools across all Kubernetes objects. For example, it finds 713 in Pods, 466 in NetworkPolicies, and 1,436 in Deployments. This shows the model can find more issues, including complex and context-dependent problems that rule-based tools miss. The model’s strength is its ability to understand different configurations and how components interact.

Traditional tools like Kube Score and Checkov do relatively well on NetworkPolicies and Roles/ClusterRoles, but our model performs much better, especially on dynamic objects like Deployments and Pods that change often and have complex settings. The high detection rates suggest the model is good at finding subtle but important misconfigurations, which can improve Kubernetes security and reliability.

\section{Threats to validity}
Several measures were taken to address potential threats to internal validity in this study of Kubernetes misconfigurations. These included thorough code reviews, rigorous testing, and multiple experimental iterations to ensure the accuracy and reliability of results. To improve external validity and generalizability, datasets were collected from diverse sources such as GitHub and Stack Overflow, covering a wide range of Kubernetes configuration files and misconfiguration examples. While ecosystem and dataset variations may affect the scope of applicability, the data diversity supports broader relevance of the findings.

\section{conclusion}
This research presents a detailed taxonomy of Kubernetes misconfigurations to improve their understanding and management in cloud-native environments. It identifies the most misconfigured Kubernetes objects that pose significant risks. The study shows that LLMs outperform traditional rule-based methods in detecting misconfigurations by capturing complex patterns. A comparison of current detection tools reveals wide variations in effectiveness, underscoring the need for advanced approaches like LLMs to strengthen cloud-native security. Future work will focus on using LLMs for both detection and automated remediation, studying misconfiguration evolution across Kubernetes versions, and developing a standardized benchmark to fairly evaluate detection and prevention tools.

\bibliographystyle{ieeetr}

\vspace{12pt}

\end{document}